# Training Process Reduction Based On Potential Weights Linear Analysis To Accelarate Back Propagation Network

Roya Asadi[1], Norwati Mustapha[2], Nasir Sulaiman[3]

[1,2,3]Faculty of Computer Science and Information Technology,
University Putra Malaysia, 43400 Serdang, Selangor, Malaysia.

[1]royaasadi@yahoo.com, [2,3]{norwati, nasir}@fsktm.upm.edu.my)

*Abstract*—**Learning is the important property of Back Propagation Network (BPN) and finding the suitable weights and thresholds during training in order to improve training time as well as achieve high accuracy. Currently, data pre-processing such as dimension reduction input values and pre-training are the contributing factors in developing efficient techniques for reducing training time with high accuracy and initialization of the weights is the important issue which is random and creates paradox, and leads to low accuracy with high training time. One good data preprocessing technique for accelerating BPN classification is dimension reduction technique but it has problem of missing data. In this paper, we study current pre-training techniques and new preprocessing technique called Potential Weight Linear Analysis (PWLA) which combines normalization, dimension reduction input values and pre-training. In PWLA, the first data preprocessing is performed for generating normalized input values and then applying then by pre-training technique in order to obtain the potential weights. After these phases, dimension of input values matrix will be reduced by using real potential weights. For experiment results XOR problem and three datasets, which are SPECT Heart, SPECTF Heart and Liver disorders (BUPA) will be evaluated. Our results, however, will show that the new technique of PWLA will change BPN to new Supervised Multi Layer Feed Forward Neural Network (SMFFNN) model with high accuracy in one epoch without training cycle. Also PWLA will be able to have power of non linear supervised and unsupervised dimension reduction property for applying by other supervised multi layer feed forward neural network model in future work.**

## I. INTRODUCTION

Back propagation network [35] uses a nonlinear supervised learning algorithm, which uses data to adjust the network's weights and thresholds for minimizing the error in its predictions on the training set. Training of BPN is considerably slow because biases and weights have to be updated in hidden layers each epoch of learning by weighted functions [26]. In Supervised Multi Layer Neural Network (SMNN) model, suitable data pre-processing techniques are necessary to find input values while pre-training techniques to find desirable weights that in turn will reduce the training process. Without preprocessing, the classification process will be very slow and it may not even complete [12]. Currently, data preprocessing specially dimension reduction and pre-training are main ideas in developing efficient techniques for fast BPN, high accuracy and reducing the training process [34, 14], but the problems in finding the suitable input values, weights and thresholds without using any random value are still remain [5, 1, 18, 12]. The current pre-training techniques generate suitable weights for reducing the training process but applying random values for initial weights [34, 4] and this will create paradox [39, 7]. This paper is organized as follows: To discuss and survey some related works in preprocessing techniques used in BPN and to compare them with new technique of PWLA. New preprocessing technique of PWLA is combination of normalization, dimension reduction and pre-training in BPN results in worthy input values, desirable process, and higher performance in both speed and accuracy. Finally, the experimental results and conclusion with future works are reported respectively.

## II. EXISTING TECHNIQUES FOR PREPROCESSING

Pre-processing in real world environment focuses on data transformation, data reduction, and pre-training. Data transformation and normalization are two important aspects of pre-processing. Transformation data is codification of values of each row in input values matrix and changing them to one data. Data transformation often uses algebraic or statistical formulas. Data normalization data transforms one input value to another suitable data by distributing and scaling. Data reduction such as dimension reduction and data compression are applied minimizing the loss of information content. SMNN models such as Back-propagation Network (BPN) are able to identify information of each input based on its weight, hence increasing the processing speed [34]. Pre-training techniques reduce the training process through preparation of suitable weights. This is an active area of research in finding efficient technique of data pre-processing for fast back-propagation network with high accuracy [5, 1, 18, 12]. Currently, data pre-processing and pre-training are the contributing factors in developing efficient techniques for fast SMNN processing at high accuracy and reduced training time [34, 14].

### A. Data Preprocessing

Several powerful data pre-processing functions on ordinal basis of improving efficiency will be discussed as latest methods in this study. These are mathematical and statistical functions to scale, filter, and pre-process the data. Changing the input data or initial conditions can immediately affect the classification accuracy in back-propagation network [5] and will be discussed as current methods in this study.

#### 1) MinMax as preprocessing:

Neal et al. explained about predicting the gold market including an experiment on scaling the input data [28]. MinMax technique will be used in BPN to transform and to scale the input values between 0 and 1 if the activation function is used the standard sigmoid and -1 to 1 for accelerating process [12, 31]. The technique of using Log (input value) is similar to MinMax for range [0..1). Another similar method is Sin( Radian(input value) ) between -π to π where Radian(input value) be between 0 to π [5]. Disadvantage of MinMax technique is lack of one special and unique class for each data [23].

#### 2) Dimension data reduction:

Dimension data reduction method projects high dimensional data matrix to lower dimensional sub-matrix for effective data preprocessing. There are two types of reduction, which are supervised and unsupervised dimension reduction. The type of reduction is based on relationship of the dimension reduction to the dataset itself or to an integrated known feature. In supervised dimension reduction, suitable sub-matrix selects based on their scores, prediction accuracy, selection the number of necessary attributes, and computing the weights with a supervised classification model. Unsupervised dimension reduction maps high dimension matrix to lower dimension and creates new low dimension matrix considering just the data points. Dimension reduction techniques are also divided into linear and nonlinear methods based on consideration of various relations between parameters. In real world, data is non-linear; hence only nonlinear techniques are able handle them. Linear techniques consider linear subset of the high dimensional space, while nonlinear techniques assume more complex subset of the high dimensional space [34, 14]. We consider Principal Component Analysis (PCA) [17] because of its properties which will be explained in pre-training section and it is known as the best dimension reduction technique until now [34, 32, 3, 24]. PCA is a classical multivariate data analysis method that is used in linear feature extraction and data compression [34, 32, 3, 24]. If the dimension of the input vectors be large, the components of the vectors are highly correlated (redundant). In this situation to reduce the dimension of the input vectors is useful. The assumption is most information in classification of high dimensional matrix has large variation. In PCA often are computed maximizing the variance in process environment for standardized linear process. The disadvantage of PCA is not to be able to find non linear relationship within input values; therefore these data will be lost. Linear Discriminant Analysis (LDA) is one dimension reduction technique based on solution of the eigenvalue problem on the product of scatter matrixes [9]. LDA computes maximizing the ratio of between-class distance to withinclass distance and sometimes singularity. LDA has three majors based on this singularity problem: regularized LDA, PCA+LDA [2] and LDA/GSVD [36, 16]. 2003). These techniques use Singular Value Decomposition (SVD) [19] or Generalized Singular Value Decomposition (GSVD). QR is one of dimensional reduction techniques for solving standard eigen problems in general matrix [11]. QR reduces the matrix to quasi triangular by unitary similarity transformation. The time complexity of QR is much smaller than SVD method [33]. Another technique is LDA/QR which maximizes the separability between different classes by using QR Decomposition [37]. The main disadvantage of PCA and other dimension reduction techniques is missing input values.

### B. Pre-training

Initialization of weights is the first critical step in training processing back-propagation networks (BPN). Training of BPN can be accelerated through a good initialization of weights during pre-training. To date, random numbers are used to initialize the weights [34, 4]. The number of epochs in training process depends on initial weights. In BPN, correct weights results in successful training. Otherwise, BPN may not obtain acceptable range of results and may halt during training. Training of BPN includes activation function in hidden layers for computing weights. Usually, initializations the weights in pre-training of BPN are random. Most papers do not report evaluation of speed and accuracy, only some comments about initializing of weights, network topology such as the number of layers and unknown practical nodes, if any. In turn, processing time depends on initial values of weights and biases, learning rate, as well as network topology [39, 7]. In the following sections, latest methods of pre-training for BPN are discussed.

#### 1) MinMax:

There are several initial weights methods which are current methods even now such as MinMax [39, 7]. In the methods of MinMax, initial weights is considered in domain of (-a, +a) which computed experimentally. SBPN is initialed random weights in domain [-0.05, 0.05]. There is an idea that initializing with large weights are important [15]. The input values are classified in three groups, the weights of the most important inputs initialized in [0.5, 1], the least important initialized with [0, 0.5] and the rest initialized with [0, 1]. The first two groups contain about one quarter of the total number input values and the other group about one half. Another good idea was introduced for initializing weight range in domain of [−0.77, 0.77] with fixed variance of 0.2 and obtained the best mean performance for multi layer perceptrons with one hidden layer [20]. The disadvantage of the method of MinMax is usage of initialization with random numbers which create critical in training.

### 2) SCAWI:

The method called Statistically Controlled Activation Weight Initialization (SCAWI) was introduced in [6]. They used the meaning of paralyzed neuron percentage (PNP) and concepted on testing how many times a neuron is in a completed situation with acceptable error. The formula of $W_{ij}^{input} = 1.3 / (1+N\ input.V^2)^{1/2}$. $r_{ij}$ is for initializing weights W that V is the mean squared value of the inputs and $r_{ij}$ is a random number uniformly distributed in the range [-1, +1]. This method was improved to $W_{ij}^{hidden} = 1.3 / (1+0.3.\ N\ hidden)^{1/2}$ . $r_{ij}$ for earning better result [8, 10]. The disadvantage of the method of SCAWI is using random numbers to put the formula and it is similar to MinMax method which has critical in training.

### 3) Multilayer auto-encoder networks as pre-training:

Codification is one of five tasks types in neural network applications [13]. Multilayer encoders are feed-forward neural networks for training with odd number of hidden layers [34, 4]. The feed forward neural network trains to minimize the mean squared error between the input and output by using sigmoid function. High dimension matrix can reduce to the low-dimensional by extracting the node values in the middle hidden layer. In addition Auto-encoder/Auto-Associative neural networks are neural networks which are trained to recall their inputs. When the neural network uses linear activation functions, auto-encoder processes are similar to PCA [22]. Sigmoid activation allows to the auto-encoder network to train a nonlinear mapping between the high-dimensional and low-dimensional data matrix. After the pre-training phase, the model is called "unfolded" to encode and decode that initially use the same weights. BPN can use for global fine-tuning phase through the whole auto-encoder to fine-tune the weights for optimization. For high number of multilayer auto-encoders connections BPN approaches are quite slowly. The auto-encoder network is unrolled and is fine tuned by a supervised model of BPN in the standard way. Since the 1980s, BPN has been obvious by deep auto-encoders and the initial weights were close enough to a good result. An auto-encoder processes very similar to PCA [22]. The main disadvantage of this method is due to the high number of multi-layer auto-encoders connections in BPN training process, resulting in slow performance.

## III. POTENTIAL WEIGHTS LINEAR ANALYSIS (PWLA)

In this study, we will compare current preprocessing techniques of BPN with new technique of Potential Weights Linear Analysis (PWLA). PWLA reduces training process of supervised multi layer neural network models. SMNN models such as BPN will change to new SMFFNN models by using real weights [30]. PWLA model recognizes high deviations of input values matrix from global mean similar to PCA and using the meaning of vector torque formula for new SMFFNN.

These deviations cause more scores for their values. For data analyzing, the first PWLA normalizes input values as data preprocessing and then uses normalized values for pre-training, at last reduces dimension of normalized input values by using their potential weights. Figure 1 illustrates the structure of Potential weight Linear Analyze preprocessing.

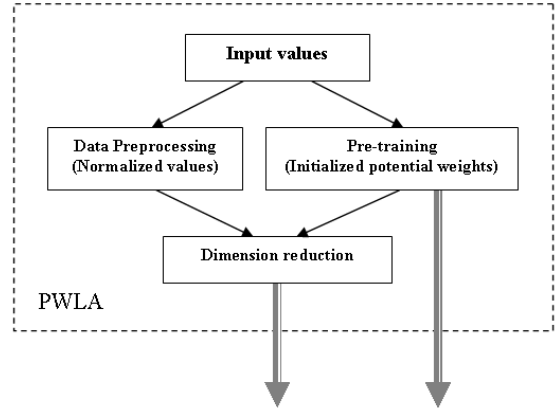

Figure 1.   Structure of Potential weight Linear Analyze

Each normalized value vector creates one vector torque ratio to global mean of matrix. They are evaluated together and they will reach to equilibrium. Figure 2 shows the example of the action for four vector torques, but all vectors create their own vector torques:

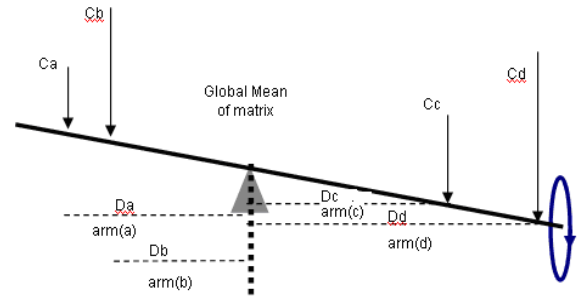

Figure 2.   The action for four vector torques

$C_a$, $C_b$, $C_c$, $C_d$ are the vectors of values while $D_a$, $D_b$, $D_c$, $D_d$ are the arms of vector torques of values. These arms are based on their distances from global mean point of matrix. The vector torque of $C_a$ is $C_a \times D_a$, the vector torque of $C_b$ is $C_b \times D_b$, the vector torque of $C_c$ is $C_c \times D_c$ and the vector torque of $C_d$ is $C_d \times D_d$. In this study, the physical and mathematical meaning of vector torque is used, and is used in classification of instances. If two input vectors have much correlation together, they will create noise. When correlation is 1, this means that the two attributes are indeed one attribute and duplication exists in input values matrix. But the global mean moves to new location, hence after equivalence, the global mean take place as one constant at special point of axis of vector torques and the weight will distribute between two vectors. This situation can be between some or all input vectors of matrix. Therefore PWLA can solve noise problem. After illustration of weak and strong weights (arms), PWLA can omit weak weights and just enter strong weights to training process of SMFFNN model. This phase called dimension reduction of normalized values in PWLA. PWLA will be more effective by having dimension reduction technique.

### A. Phases of Weights Linear Analysis

The input values can be every numeric type, range and measurement unit. Table I shows the input values matrix:

TABLE I.    INPUT VALUES MATRIX

|  | Attribute $_1$ | Attribute $_2$ |  | Attribute $_m$ |
|---|---|---|---|---|
| Instance $_1$ | $X_{11}$ | $X_{12}$ | ... | $X_{1m}$ |
| Instance $_2$ | $X_{21}$ | $X_{22}$ | ... | $X_{2m}$ |
| ... | ... | ... | ... | ... |
| Instance $_n$ | $X_{n1}$ | $X_{n2}$ | ... | $X_{nm}$ |

If the dataset is large, there a high chance that the vector components are highly correlated (redundant). Even though the input values have correlation, hence the input is noisy; PWLA method is able to solve this problem. The covariance function returns the average of deviation products for each data value pair two attributes. There are three phases in implementing PWLA:

### 1) Normalization:

In this phase, normalized values as data pre-processing is considered. The technique of Min and Max is used. Each value is computed to find the ratio to average of all columns. Table II shows a rational distribution of information in each row. The table is used as normalized input values or input vectors.

TABLE II.    NORMALIZING INPUT VALUES PHASE

|  | Attribute $_1$ | Attribute $_2$ | ... | Attribute $_m$ |
|---|---|---|---|---|
| Instance $_1$ | $C_{11}=X_{11}/$ $Ave_1$ | $C_{12}=X_{11}/Ave_2$ | ... | $C_{1m}=X_{1m}/$ $Ave_m$ |
| Instance $_2$ | $C_{21}=X_{21}/$ $Ave_1$ | $C_{22}=X_{22}/Ave_2$ | ... | $C_{2m}=X_{2m}/$ $Ave_m$ |
| ... | ... | ... | ... | ... |
| Instance $_n$ | $C_{n1}=X_{n1}/$ $Ave_1$ | $C_{n2}=X_{n2}/Ave_2$ | ... | $C_{nm}=X_{nm}/$ $Ave_m$ |
| Average | $Ave_1$ | $Ave_2$ | $Ave_{...}$ | $Ave_m$ |

### 2) Pre-training:

In improving pre-training performance, potential weights are initialized. The first distribution of standard normalized values is computed. $\mu$ is mean of values vectors each row, and $\sigma$ is standard deviation of values vectors each row. $Z_{nm}$ is a standard normalized value and is computed based on formula below.

$$Z_{nm} = ( X_{nm} - \mu_m ) / \sigma_m$$

Therefore, PWLA does not need to use any random number for initialization of potential real weights and input of pre-training phase is normalized input values. Table III shows horizontal evaluation of PWLA that is about pre-training.

TABLE III.    HORIZONTAL EVALUATION OF PWLA

|  | Attribute $_1$ | Attribute $_2$ | .. | Attribute $_m$ | $\mu$ | $\Sigma$ |
|---|---|---|---|---|---|---|
| Instance $_1$ | $Z_{11}$ | $Z_{12}$ | .. | $Z_{1m}$ | $\mu_1$ | $\Sigma_1$ |
| Instance $_2$ | $Z_{21}$ | $Z_{22}$ | .. | $Z_{2m}$ | $\mu_2$ | $\Sigma_2$ |
| ... | ... | ... | .. | ... | ... | ... |
| Instance $_n$ | $Z_{n1}$ | $Z_{n2}$ | .. | $Z_{nm}$ | $\mu_m$ | $\Sigma_m$ |

We explained the arms of value vectors are computed based on definition of deviation and distribution of standard normalization. $Z_{nm}$ shows the distance of $C_{nm}$ to mean its rows. Global mean is the center of vectors torques. The weights are arms in vectors torques. This definition of weight is based on statistical and mathematical definition of normalization distribution and vector torque. $W_m$ is equivalent to ( $|Z_{11}|+|Z_{22}|+|Z_{...}|+|Z_{nm}|$ ) / $n$. $|Z_{nm}|$ is absolute of normal value $Z_{nm}$. Hence, weight selection is not randomization. The weights may have thresholds but must be managed in hidden layer of new SMFFNN using the following equation.

$$W_m = ( |Z_{11}|+|Z_{22}|+|Z_{...}|+|Z_{nm}| ) / n$$

### 3) Dimension reduction:

In this phase, there are normalized values and potential real weights. The weights show deviations of input values matrix from global mean similar to PCA. PWLA with having potential real weights can recognize high dimensional data matrix for effective data preprocessing. The suitable sub-matrix of necessary attributes can be selected based on their potential weights. PWLA can map high dimension matrix to lower dimension matrix. The strong weight causes high variance. If the dimension of the input vectors be large, the components of the vectors are highly correlated (redundant). PWLA can solve this problem in two ways. The first, after equivalence, the global mean take place as one constant at special point of axis of vector torques and the weights are distributed between vectors. In other way, it can solve redundancy by dimension reduction. This phase of PWLA can be performed in hidden layer during pruning.

### B. PWLA Algorithm:

The algorithm of PWLA is shown in Figure 3.

PWLA (D; L, W)
Input: Database D, database of input values;
Output: Matrix L, Normalized database of D; W, potential weights;
Begin
//Computing vertical evaluation: In this phase, the input values are translated.
Let row number: n;
Let column number: m;
Let copy of database D in Matrix n×m of L;

*Forall columns of Matrix L m do*
    *Forall rows of Matrix L n do   {*
    *L(n,m)= L(n,m)/Average(column m);*
    *Matrix LTemp = copy of Matrix L;  }*
*//Computing horizontal evaluation: In second phase of procedure, the weight of each input value is computed.*
*//Computing standard normalized values in row:*
*Let $\mu_n$ = Mean of values vectors in row;*
*Let $\sigma_n$ = Standard deviation of values vectors in row;*
*Forall columns of Matrix L m do*
    *Forall rows of Matrix L n do*
    *LTemp(n,m) = (LTemp(n,m) − $\mu_n$ ) /$\sigma_n$ ;*
*// Computing arms of values vectors (Weights):*
*Forall columns of Matrix L m do*
*$W_m$ = (Average of Absolute (LTemp( column m)) ;*
*Apply dimension reduction Matrix L*
*Return Matrix L, potential weights W*

Figure 3.   Algorithm of PWLA

The time complexity of PWLA technique depends on the number of attributes p and the number of instances n. The technique of PWLA is linear and its time complexity will be O(pn). PWLA technique output dimension reduction of normalized input values and potential weights. New SMFFNN will process based on the algebraic consequence of vectors torques. The vectors torque $T_{nm} = C_{nm} \times W_m$ are the basis of the physical meaning of torque. Each torque $T$ shows a real worth of each value between whole values in matrix. The algebraic consequence of vectors torques $\check{S}_n$ is equivalent to $T_{n1} + T_{n2} + T + \ldots + T_{nm}$. In each row, $\check{S}_n$ is computed. The output will be classified based on $\check{S}_i$. Recall that BPN uses sigmoid activation function to transform actual output between domain [0, 1] and to compute error by using the derivative of the logistic function in order to compare actual output with true output. True output forms the basis of the class labels in a given training dataset.

Here, PWLA computes potential weights and new SMFFNN computes desired output by using binary step function instead of sigmoid function as activation function. Also there is no need to compute error and the derivative of the logistic function for the purpose of comparison between actual output with true output. The output $\check{S}_n$ are sorted and two stacks are created based on true output with class label 0 and class label 1. Thresholds are defined based on creation of Stack0 as Stack of $I_j$ with condition of class label 0, Stack1 as Stack of $I_j$ with condition of class label 1. Binary step function is applied to both stacks serving as threshold and generated desired output 0 or 1. New SMFFNN applies PWLA similar to the simple neural network. The number of layers, nodes, weights, and thresholds in new SMFFNN using PWLA pre-processing is logically clear without presence of any random elements. New SMFFNN will classify input data by using output of PWLA, whereby there is one input layer with several input nodes, one hidden layer, and one output layer with one node. Hidden layer contains of weighted function $\sum_i W_{ij} I_i$ and $W_{ij}$ are potential weights from pre-training. Hidden layer is necessary for pruning or considering management opinions for weights optimization. In pruning, the input values with weak weights can be omitted because they have weak effect on desired output. In management strategies, the input values with weights bigger than middle weights perform effectively on desired output, therefore they can be optimized. The output node in output layer contains $\sum_j W_{jo} I_j$ and $W_{jo}$=1 for computing desired output. Here, BPN exist only in one epoch during training processing without the need to compute bias and error in hidden layer. In evaluating the test set and predicting the class label, weights and thresholds are clear and class label of each instance can be predicted by binary step function.

## IV.   Exprimental Results and Discussion

All techniques have been implemented in Visual Basic version 6 and all performed on 1.662 GHz Pentium PC with 1.536 GB of memory. Back-propagation network performed with configuration of 10 hidden units in one hidden layer, and 1 output unit. We consider initial random weights in range of [-0.77, 0.77] for standard BPN in experiment results. We considered F-measure or balanced F-Score to compute average of accuracies test on 10 folds across [12]. The variables of F-measure are as follow and based on weighting of recall and precision. Recall is the probability that a randomly selected relevant instance is recovered in a search, Precision is the probability that a randomly selected recovered instance is relevant.

$t_p$= true positive    ;    $f_p$= false positive
$t_n$= true negative    ;    $f_n$= false negative
Recall= $t_p$ / $t_p$+$f_n$    ;    Precision= $t_p$ / $t_p$+$f_p$
F= 2. (Precision. recall) / (precision+ recall)

### A)  XOR problem:

To illustrate the semantic and logic of the proposed technique and the new SMFFNN model, the problem of Exclusive-OR (XOR) is considered.

$$X_1 \oplus X_2 = \text{XOR}(X)$$

There are two logical attributes and four instances. Attribute characteristic is binary 0, 1. Usually XOR is used by multi-layers artificial neural networks. The analysis of XOR problem is illustrated in Table IV, together with its features and class label.

TABLE IV.    The Features of XOR

|  | Attribute 1 | Attribute 2 | Class label of XOR |
|---|---|---|---|
| **Instance 1** | 0 | 0 | 0 |
| **Instance 2** | 0 | 1 | 1 |
| **Instance 3** | 1 | 0 | 1 |
| **Instance 4** | 1 | 1 | 0 |

Learning of new SMFFNN using PWLA takes one epoch without computing sigmoid function, training cycle, mean square error, and updating weights. Potential weights are obtain through PWLA and new SMFFNN model applies all them to compute thresholds and binary step function for generating desired output as well as predicting class label for XOR. The potential weights of attribute1 and attribute2 are the same (0.5) by using PWLA because the correlation between the values of attribute1 and attribute 2 is zero ($\rho = 0$). In this case, the output of new model shows the error is 0 and the outputs are the same class labels. Figure 4 illustrates result of new SMFFNN by using PWLA implementation.

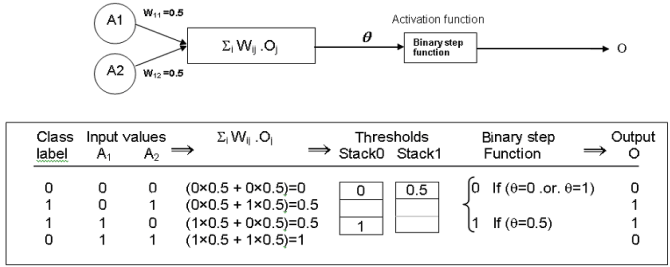

Figure 4. Implementation of XOR problem by new SMFFNN using PWLA

Training results are shown in The problem of XOR is implemented using PWLA and the new SMFFNN. The experimental results are compared with results from Standard BPN and Improved BPN [25, 38]. Table V shows comparison of speed and accuracy of XOR problem.

TABLE V. CLASSIFICATION OF THE XOR PROBLEM

| Classification model | Number of epoch | error |
|---|---|---|
| New SMFFNN with PWLA | 1 | - |
| Improved BPN | 3167 | 0.0001 |
| Standard BPN | 7678 | 0.0001 |
| PCA+BPN | 200 | 0.0002 |

The result of New SMFFNN with PWLA is better than others.

*B) SPECT Heart:*

SPECT Heart is selected from UCI Irvine Machine Learning Database Repository [27] because the implementations of the neural network models on this dataset was remarkable since most conventional methods do not process well on these datasets [21]. The dataset of SPECT Heart contains diagnosing of cardiac Single Proton Emission Computed Tomography images. There are two classes: normal and abnormal. The database contains 267 SPECT image sets of patients features, 22 continuous feature patterns. The implementation of BPN and new SMFFNN models by using different preprocessing techniques are compared. The learning process of new SMFFNN using PWLA is performed in one epoch and the potential weights are generated by PWLA on SPECT Heart training dataset which are shown in Table VI.

TABLE VI. POTENTIAL WEIGHTS OF SPECT HEART (TRAINING DATASET)

| Attribute | 1 | 2 | 3 | 4 | 5 | 6 | 7 | 8 | 9 | 10 | 11 |
|---|---|---|---|---|---|---|---|---|---|---|---|
| Weight | 0.039 | 0.044 | 0.044 | 0.049 | 0.045 | 0.052 | 0.041 | 0.044 | 0.045 | 0.044 | 0.047 |
| Attribute | 12 | 13 | 14 | 15 | 16 | 17 | 18 | 19 | 20 | 21 | 22 |
| Weight | 0.040 | 0.042 | 0.043 | 0.053 | 0.041 | 0.046 | 0.047 | 0.047 | 0.051 | 0.043 | 0.054 |

New SMFFNN considered the potential weights and obtained thresholds and then created two stacks for 0 and 1 labels using binary step function. The potential weights of SPECT Heart have suitable distribution and are near together. Table VII shows stacks of thresholds for SPECT Heart by using new SMFFNN and PWLA.

TABLE VII. CREATED THRESHOLDS BY NEW SMFFNN USING PWLA ON SPECT HEART (TRAINING SET)

| Stack0 | Stack0 | Stack1 | Stack1 |
|---|---|---|---|
| 0.031529 | 2.84E-03 | 5.00E-02 | 1.01E-02 |
| 2.15E-02 | 2.06E-03 | 4.84E-02 | 9.50E-03 |
| 2.04E-02 | 1.51E-03 | 4.27E-02 | 9.08E-03 |
| 1.83E-02 |  | 4.15E-02 | 8.26E-03 |
|  | 0 | 3.76E-02 |  |
| 1.24E-02 |  | 3.40E-02 | 7.68E-03 |
| 1.14E-02 |  | 3.11E-02 | 7.12E-03 |
| 1.13E-02 |  | 2.92E-02 |  |
| 9.73E-03 |  | 2.60E-02 | 4.13E-03 |
|  |  | 2.46E-02 |  |
| 8.26E-03 |  | 2.33E-02 | 3.41E-03 |
| 7.93E-03 |  | 2.25E-02 |  |
| 7.68E-03 |  | 1.96E-02 | 2.84E-03 |
| 7.22E-03 |  | 1.95E-02 | 2.06E-03 |
|  |  | 0.018773 |  |
| 6.97E-03 |  | 0.018336 |  |
| 6.47E-03 |  | 1.79E-02 |  |
| 6.17E-03 |  | 1.66E-02 |  |
| 6.94E-03 |  | 1.62E-02 |  |
| 5.68E-03 |  | 1.51E-02 |  |
| 5.16E-03 |  | 1.39E-02 |  |
|  |  | 1.31E-02 |  |
| 4.11E-03 |  | 1.19E-02 |  |
| 3.72E-03 |  | 1.11E-02 |  |
|  |  | 1.08E-02 |  |
| 3.25E-03 |  | 1.02E-02 |  |
| 2.86E-03 |  |  |  |

The learning of BPN is performed in standard situation (SBPN), and by using PCA with 10 dimensions as preprocessing technique. Table VIII shows speed and accuracy of classification methods on SPECT Heart dataset.

TABLE VIII. COMPARISON OF CLASSIFICATION SPEED AND ACCURACY ON SPECT HEART DATASET

| The classification methods | Accuracy | Epoch | CPU time (second) |
|---|---|---|---|
| New SMFFNN by using PWLA | 92.00% | 1 | 0.036 |
| New SMFFNN by using PWLA with dimension reduction | 87% | 1 | 0.019 |
| SBPN | 87.00% | 25 | 2.92 |
| BPN by using PCA | 73.30% | 14 | 1.08 |

In SPECT Heart, The accuracy of BPN by using PCA is 73.3%, SBPN is 87%. New SMFFNN by using PWLA has higher accuracy to others which is 92% and by using PWLA with dimension reduction is 87%. We consider 11 attributes with high weights for dimension reduction of input values matrix. Figure 5 shows the chart of comparison accuracy of classification methods on SPECT Heart dataset.

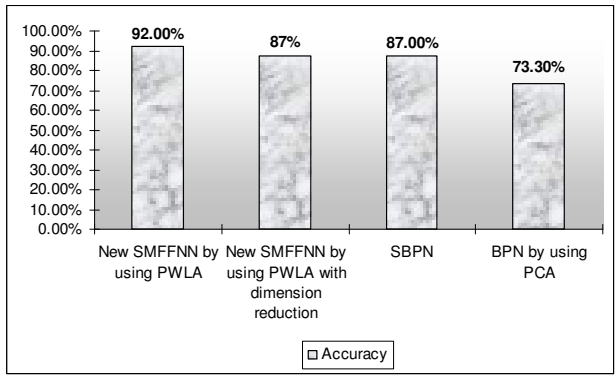

Figure 5. Comparison of classification accuracy on SPECT Heart dataset

The accuracy of performance of new SMFFNN by using PWLA is better than other methods because it uses real potential weights, thresholds and does not work on random initialization. SBPN method performs on SPECT Heart training dataset in 25 epochs with 2.92 second CPU times. BPN by using PCA performs on SPECT Heart training dataset in 14 epochs with 1.08 second CPU. New SMFFNN by using PWLA in one epoch during 0.036 second processes on SPECTF Heart training dataset. New SMFFNN by using PWLA with dimension reduction in one epoch during 0.019 second processes on SPECTF Heart training dataset which has higher speed to other methods. Comparison on speed of training process of the methods on SPECT Heart dataset is shown in Figure 6 as follow:

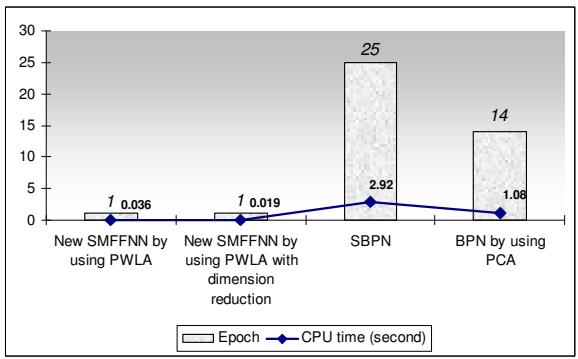

Figure 6. Comparison of classification speed of BPN and new SMFFNN by using preprocessing techniques on SPECT Heart dataset

## C) SPECTF Heart:

SPECTF Heart is selected from UCI Irvine Machine Learning Database Repository [27] because the implementations of the neural network models on these datasets were remarkable since most conventional methods do not process well on these datasets [21]. The dataset of SPECTF Heart contains diagnosing of cardiac Single Proton Emission Computed Tomography (SPECT) images. There are two classes: normal and abnormal. The database contains 267 SPECT image sets of patients features, 44 continuous feature patterns.

The implementation of BPN and new SMFFNN models by using different preprocessing techniques are compared.

Learning for the new SMFFNN using PWLA is performed in only one epoch. Generated potential weights by PWLA on SPECTF HEART training dataset are shown in Table IX.

TABLE IX.    POTENTIAL WEIGHTS OF SPECTF HEART (TRAINING DATASET)

| Attribute | 1 | 2 | 3 | 4 | 5 | 6 | 7 | 8 | 9 | 10 |
|---|---|---|---|---|---|---|---|---|---|---|
| Weight | 2.422 | 2.697 | 2.055 | 2.004 | 2.227 | 2.006 | 1.741 | 1.931 | 2.205 | 2.458 |
| Attribute | 11 | 12 | 13 | 14 | 15 | 16 | 17 | 18 | 19 | 20 |
| Weight | 2.106 | 2.176 | 1.847 | 1.825 | 2.132 | 2.295 | 1.879 | 2.094 | 2.47 | 2.399 |
| Attribute | 21 | 22 | 23 | 24 | 25 | 26 | 27 | 28 | 29 | 30 |
| Weight | 1.930 | 1.710 | 2.038 | 2.197 | 3.157 | 3.630 | 2.869 | 2.977 | 2.125 | 2.4 |
| Attribute | 31 | 32 | 33 | 34 | 35 | 36 | 37 | 38 | 39 | 40 |
| Weight | 1.476 | 1.373 | 1.819 | 1.960 | 1.740 | 1.710 | 2.524 | 2.809 | 2.071 | 2.148 |
| Attribute | 41 | 42 | 43 | 44 | | | | | | |
| Weight | 2.700 | 2.823 | 3.169 | 3.686 | | | | | | |

The potential weights don't have suitable distribution; therefore we can consider dimension reduction of input values based on weak weights. In here, we consider 14 attributes with high weights for dimension reduction technique. New SMFFNN considered these potential weights and obtained thresholds before it created two stacks for 0 and 1 labels using binary step function. Table X shows stacks of thresholds.

TABLE X.    CREATED THRESHOLDS BY NEW SMFFNN USING PWLA ON SPECTF HEART (TRAINING SET)

| Stack0 | Stack0 | Stack1 | Stack1 |
|---|---|---|---|
| 1.341693 | 1.276903 | 1.319296 | 1.228601 |
| 1.332656 | 1.274138 | 1.309951 | |
| 1.331425 | | | 1.225405 |
| 1.327727 | 1.2704 | 1.296126 | 1.22212 |
| 1.321062 | 1.268604 | 1.293849 | 1.221851 |
| 1.320334 | | | 1.213224 |
| | 1.267201 | | 1.206418 |
| 1.318787 | 1.264989 | 1.293434 | 1.20177 |
| 1.317596 | 1.264238 | | 1.199884 |
| 1.317446 | 1.263424 | 1.292092 | 1.172046 |
| 1.316756 | | 1.292003 | 1.166529 |
| 1.314495 | 1.25403 | | 1.164348 |
| | | 1.289961 | 1.129222 |
| 1.309789 | 1.248756 | | 1.107869 |
| 1.30963 | | 1.272229 | 1.098162 |
| 1.308203 | 1.236021 | 1.27151 | 1.022269 |
| 1.307151 | | 1.270787 | 1.019312 |
| 1.304646 | 1.229239 | | 0.969073 |
| 1.301085 | | 1.26759 | 0.866963 |
| 1.300473 | 1.225636 | | |
| 1.299745 | | 1.263384 | |
| | | 1.259336 | |
| 1.293854 | | 1.258056 | |
| 1.292669 | | | |
| | | 1.251074 | |
| 1.291389 | | 1.246279 | |
| | | 1.240893 | |
| 1.28993 | | 1.239723 | |
| 1.285006 | | 1.238957 | |
| 1.279928 | | 1.236642 | |
| 1.277346 | | | |
| 1.277079 | | 1.230332 | |

The learning of BPN is performed in standard situation (SBPN), and by using PCA with 10 dimensions as preprocessing technique. Table XI shows speed and accuracy of classification methods on SPECTF Heart dataset.

TABLE XI.   COMPARISON OF CLASSIFICATION SPEED AND ACCURACY ON SPECTF HEART DATASET

| The classification methods | Accuracy | Epoch | CPU time (second) |
|---|---|---|---|
| New SMFFNN by using PWLA | 94.00% | 1 | 0.061 |
| New SMFFNN by using PWLA with dimension reduction | 85% | 1 | 0.022 |
| SBPN | 79.00% | 25 | 4.98 |
| BPN by using PCA | 75.10% | 14 | 1.6 |

In SPECTF Heart, the accuracy of BPN by using PCA is 75.1%, SBPN is 79% and new SMFFNN by using PWLA has higher accuracy to others which is 94% and by using PWLA with dimension reduction is 85%. We consider 14 attributes with high weights for dimension reduction of input values matrix. Figure 7 shows the charts of comparison accuracy of classification methods on SPECTF Heart dataset.

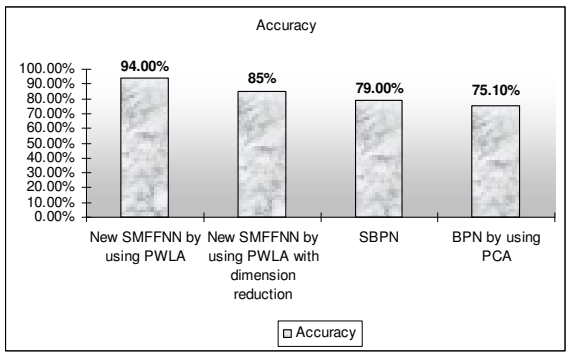

Figure 7.   Comparison of classification accuracy on SPECTF Heart dataset

The accuracy of performance of new SMFFNN by using PWLA is better than other methods because it uses real potential weights, thresholds and does not work on random initialization. SBPN method performs on SPECTF Heart training dataset in 25 epochs with 4.98 second CPU times. BPN by using PCA performs on SPECTF Heart training dataset in 14 epochs with 1.6 second CPU times. New SMFFNN by using PWLA in one epoch during 0.061 second processes on SPECTF Heart training dataset. New SMFFNN by using PWLA with dimension reduction in one epoch during 0.022 second processes on SPECTF Heart training dataset which has higher speed to other methods. Speed comparison of models and techniques on SPECTF Heart dataset are shown in Figure 8.

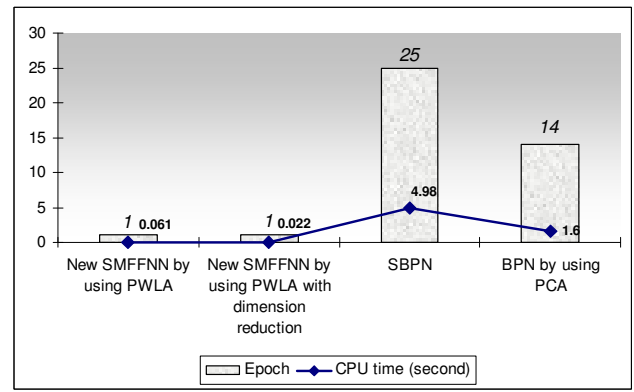

Figure 8.   Comparison of classification speed of BPN and new SMFFNN by using preprocessing techniques on SPECTF Heart dataset

### D)   Liver disorders dataset (BUPA):

Liver disorders dataset or BUPA Medical Research Ltd. database donated by Richard S. Forsyth is selected from UCI Irvine Machine Learning Database Repository [27]. The first 5 variables are all blood tests which are thought to be sensitive to liver disorders that might arise from excessive alcohol consumption. Each line in the BUPA data file constitutes the record of a single male individual. Selector field used to split data into two sets. The database contains 345 attributes and 7 instances. Learning for the new SMFFNN using PWLA is performed in only one epoch and the potential weights generated by PWLA on BUPA dataset is shown in Table XII.

TABLE XII.   POTENTIAL WEIGHTS OF BUPA (TRAINING DATASET)

| Attribute | 1 | 2 | 3 | 4 | 5 | 6 |
|---|---|---|---|---|---|---|
| Weight | 14.562 | 14.879 | 16.587 | 15.770 | 19.088 | 19.115 |

New SMFFNN considered these potential weights and obtained thresholds before it created two stacks for 1 and 2 labels using binary step function. Table XIII shows stacks of thresholds.

TABLE XIII. CREATED THRESHOLDS BY NEW SMFFNN USING PWLA ON BUPA (TRAINING SET)

| Stack1 | Stack1 | Stack1 | Stack1 | Stack1 | Stack1 |
|---|---|---|---|---|---|
| 0.404405 | 0.294897 | 0.388393 | 0.674031 | 0.330233 | 0.40023 |
| 0.251772 | 0.281024 | 0.393958 | 0.509848 | 0.294946 | 0.57203 |
| 0.246235 | 0.301343 | 0.410628 | 0.484096 | 0.319875 | 0.355366 |
| 0.256148 | 0.370683 | 0.397226 | 0.6155 | 0.366758 | 0.385436 |
| 0.260036 | 0.348065 | 0.470395 | 0.6155 | 0.231955 | 0.421104 |
| 0.260281 | 0.386366 | 0.443028 | 0.826280 | 0.370761 | 0.332224 |
| 0.254191 | 0.2712 | 0.309166 | 0.819531 | 0.315207 | 0.34251 |
| 0.40813 | 0.350704 | 0.441245 | 1.043013 | 0.380763 | 0.43302 |
| 0.303101 | 0.32703 | 0.422187 | 1.474236 | 0.317508 | 0.518151 |
| 0.297372 | 0.448035 | 0.453635 | 0.28423 | 0.267792 | 0.405577 |
| 0.329718 | 0.403529 | 0.48401 | 0.249399 | 0.365753 | 0.556648 |
| 0.263874 | 0.328712 | 0.462578 | 0.322293 | 0.239836 | 0.718043 |
| 0.276977 | 0.381914 | 0.526568 | 0.229921 | 0.360883 | 0.744678 |
| 0.413385 | 0.31335 | 0.389607 | 0.303486 | 0.337263 | 0.47377 |
| 0.333786 | 0.378091 | 0.564891 | 0.253226 | 0.304343 | 0.45247 |
| 0.300581 | 0.359048 | 0.63209 | 0.326731 | 0.29191 | 0.452595 |
| 0.290876 | 0.357151 | 0.748758 | 0.314384 | 0.284736 | 1.055497 |
| 0.278095 | 0.396391 | 0.453159 | 0.296241 | 0.308529 | 0.96218 |
| 0.287088 | 0.309934 | 0.462578 | 0.294851 | 0.47697 | 1.453635 |
| 0.583259 | 0.490208 | 0.795165 | 0.259406 | 0.372645 | 0.517722 |
| 0.299956 | 0.333195 | 0.871199 | 0.477912 | 0.301495 | 0.672659 |
| 0.274434 | 0.440288 | 0.6155 | 0.360896 | 0.364246 | 0.632095 |
| 0.269244 | 0.363318 | 0.572332 | 0.63053 | 0.336279 | 0.539335 |
|  |  |  |  |  | 0.617867 |

| Stack2 | Stack2 | Stack2 | Stack2 | Stack2 | Stack2 |
|---|---|---|---|---|---|
| 0.40891 | 0.26286 | 0.32378 | 0.62393 | 0.42541 | 0.66151 |
| 0.36312 | 0.19171 | 0.36776 | 0.43968 | 0.43156 | 0.4899 |
| 0.37192 | 0.25151 | 0.36267 | 0.59097 | 0.32002 | 0.42271 |
| 0.25849 | 0.24603 | 0.40217 | 0.43674 | 0.33877 | 0.39426 |
| 0.23979 | 0.32132 | 0.34957 | 0.48618 | 1.10121 | 0.36641 |
| 0.91162 | 0.33258 | 0.55912 | 0.55638 | 0.3586 | 0.38567 |
| 0.38499 | 0.28159 | 0.4341 | 0.41025 | 0.39984 | 0.42009 |
| 0.21544 | 0.75063 | 0.42348 | 0.60004 | 0.32967 | 0.75769 |
| 0.35783 | 0.38345 | 0.39609 | 0.76262 | 0.30063 | 0.33615 |
| 0.33959 | 0.30226 | 0.48132 | 0.64094 | 0.32963 | 0.41789 |
| 0.53343 | 0.40299 | 0.34571 | 0.47772 | 0.24268 | 0.53506 |
| 0.41297 | 0.53288 | 0.47611 | 1.1919 | 0.28683 | 0.43293 |
| 0.31547 | 0.44195 | 0.56835 | 0.58637 | 0.25627 | 0.44298 |
| 0.27326 | 0.44643 | 0.35015 | 0.76722 | 0.24657 | 0.33701 |
| 0.27709 | 0.33985 | 0.38248 | 0.58757 | 0.30447 | 0.37565 |
| 0.26924 | 1.02145 | 0.75078 | 0.74846 | 0.52733 | 0.64813 |
| 0.23385 | 0.33985 | 1.00733 | 0.85808 | 0.37483 | 0.64004 |
| 0.38664 | 0.38043 | 0.37614 | 0.94759 | 0.55744 | 0.47736 |
| 0.29461 | 0.31538 | 0.36052 | 0.74087 | 0.3096 | 0.42013 |
| 0.34101 | 0.50638 | 0.41407 | 0.38067 | 0.44675 | 0.52877 |
| 0.29456 | 0.58008 | 0.42743 | 0.4546 | 0.31058 | 0.38662 |
| 0.28034 | 0.40666 | 0.658 | 0.29467 | 0.49541 | 1.11519 |
| 0.60193 | 0.41283 | 0.38718 | 0.45493 | 0.4127 | 0.48205 |
| 0.3178 | 0.37594 | 0.76738 | 0.37605 | 0.29706 | 0.35384 |
| 0.29921 | 0.47627 | 0.5312 | 0.29309 | 0.41757 | 0.37197 |
| 0.30824 | 0.51839 | 0.44396 | 0.29288 | 0.41966 | 0.45884 |
| 0.30451 | 0.47691 | 0.36819 | 0.2543 | 0.33235 | 0.5681 |
| 0.27763 | 0.43692 | 0.60994 | 0.24704 | 0.27325 | 0.48108 |
| 0.36565 | 0.38252 | 0.5914 | 0.27755 | 0.36864 | 0.7204 |
| 0.31907 | 0.41872 | 0.81532 | 0.41249 | 0.42759 | 0.54361 |
| 0.36048 | 0.88338 | 0.62867 | 0.50186 | 0.53829 | 0.56059 |
|  |  |  |  |  | 0.49781 |
|  |  |  |  |  | 0.53798 |
|  |  |  |  |  | 1.12668 |
|  |  |  |  |  | 0.41754 |

Table XIV shows average accuracy of classification methods on BUPA dataset.

TABLE XIV. COMPARISON OF CLASSIFICATION SPEED AND ACCURACY ON BUPA DATASET

| The classification methods | Accuracy | Epoch |
|---|---|---|
| New SMFFNN by using PWLA | 100% | 1 |
| SBPN | 59.40% | 1300 |
| BPN by using PCA | 63.40% | 200 |
| BPN by using SCAWI | 60.90% | 200 |

In BUPA, the accuracy of BPN by using SCAWI is 60.90% [7], BPN by using PCA is 63.40% [29], SBPN is 59.40% and BPN by using PWLA has higher accuracy to others that is 100%. Figure 9 shows the charts of comparison accuracy of classification methods on BUPA dataset.

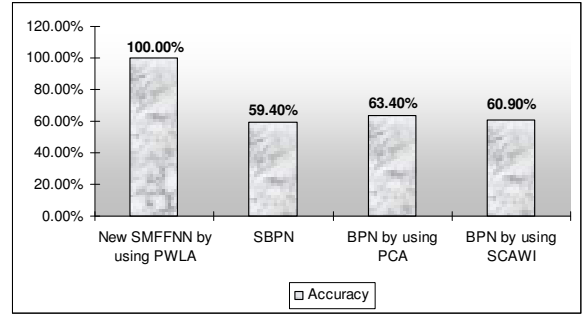

Figure 9. Comparison of classification accuracy on BUPA dataset

The accuracy of performance of new SMFFNN by using PWLA is better than other methods and is 100% because it uses real potential weights, thresholds and does not work on random initialization. The BUPA classification result shows that selected attributes of this dataset are complete and SMFFNN model by using PWLA has highest accuracy.

SBPN method performs on BUPA dataset in 1300 epochs. BPN by using PCA or SCAWI performs on BUPA dataset in 200 epochs. New SMFFNN by using PWLA in one epoch processes on BUPA dataset that has higher speed than other methods.

Speed comparison of models and techniques on BUPA dataset are shown in Figures 10.

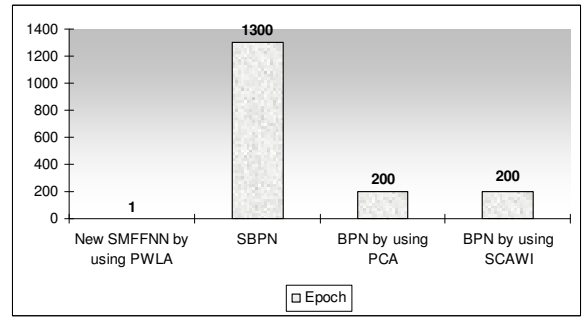

Figure 10. Comparison of classification speed of BPN and new SMFFNN by using preprocessing techniques on BUPA dataset

## V. CONCLUSION

Currently, data pre-processing and pre-training techniques of BPN focus on reducing the training process and increasing classification accuracy. The main contribution of this paper is combination of normalization and dimension reduction as pre-processing and new pre-training method for reducing training process with the discovery of SMFFNN model. BPN can change to new SMFFNN model and get the best result in speed and accuracy by using new preprocessing technique without gradient of mean square error function and updating weights in one epoch. Therefore, the proposed technique can solve the main problem of finding the suitable weights. The Exclusive-OR (XOR) problem is considered and solved for the purpose to validate the new model. During experiments, the new model was implemented and analyzed using Potential Weights Linear Analysis (PWLA). The combination of normalization, dimension reduction and new pre-training techniques shows that PWLA generated suitable input values and potential weights. This shows that PWLA serves as global mean and vectors torque formula to solve the problem. Three kinds of SPECT Heart, SPECTF Heart and Liver Disorders (BUPA) datasets from UCI Repository of Machine Learning are chosen to illustrate the strength of PWLA techniques. The results of BPN by using pre-processing techniques and new SMFFNN with application of PWLA showed significant improvement in speed and accuracy. The results show that robust and flexibility properties of new preprocessing technique for classification. For future work, we consider improved PWLA with non linear supervised and unsupervised dimension reduction property for applying by other supervised multi layer feed forward neural network models.

Department of Computer Science and Engineering, Intelligent Information Processing Laboratory, Fudan University, People's Republic of China. Accepted 31 October 2003.

**A**UTHORS PROFILE

**Dr. Prof. Md. Nasir bin Sulaiman** is a lecturer in Computer Science in Faculty of Computer Science and Information Technology, UPM and as an Associate Professor since 2002. He obtained Ph. D in Neural Network Simulation from Loughborough University, U.K. in 1994. His research interests include intelligent computing, software agents and data mining.

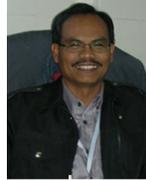

**Dr. Norwati Mustapha** is a lecturer in Computer Science in Faculty of Computer Science and Information Technology, UPM and head of department of Computer Science since 2005. She obtained Ph. D in Artificial Intelligence from UPM, Malaysia in 2005.

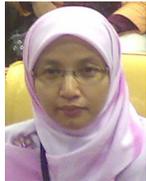

Her research interests include intelligent computing and data mining.

**Roya Asadi** received the Bachelor degree in Computer Software engineering from Electronical and Computer Engineering Faculty, Shahid Beheshti University and Computer Faculty of Data Processing Iran Co. (IBM), Tehran, Iran.

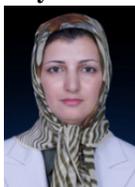

She is a research student of Master of Computer science in database systems in UPM university of Malaysia. Her professional working experience includes 12 years of service as Senior Planning Expert 1. Her interests are in Intelligent Systems and Neural Network modeling.